\documentclass{article}

\usepackage[preprint]{neurips_2026}

\usepackage[utf8]{inputenc}
\usepackage[T1]{fontenc}

\usepackage{hyperref}
\usepackage{url}

\usepackage{booktabs}
\usepackage{amsmath}
\usepackage{tcolorbox}
\tcbuselibrary{most}
\usepackage{cleveref}
\usepackage{float}

\usepackage{amssymb}

\usepackage{graphicx}
\usepackage{nicefrac}
\usepackage{microtype}
\usepackage{xcolor}
\usepackage{appendix}

\usepackage{multirow}
\usepackage{enumitem}
\usepackage{soul}

\definecolor{linkblue}{RGB}{0,82,155}

\definecolor{linkblue}{RGB}{0,92,175}

\hypersetup{
    colorlinks=true,
    linkcolor=black,      
    citecolor=linkblue,   
    urlcolor=linkblue,    
    pdfborder={0 0 0}
}

\title{The Effects of Incremental Instruction Delivery on Language-Model Creative Writing}

\author{
  Anshuman Singh \\
  SGT UNIVERSITY \\
  \texttt{anshuman.su.dev@gmail.com}
  \And
  Abrar Eyasir\thanks{Corresponding author.} \\
  CSE, University of Dhaka \\
  \texttt{eyasir2047@gmail.com}
  \And
 Haseeb Yaqoob \\
  NED University of Engineering and Technology\\
  \texttt{haseeb.yaqoob201@gmail.com}
  \And
  John Manavalan  \\
  Metea Valley High School, Illinois \\
  \texttt{professor.paranormal159@gmail.com}
}

\begin{document}
\raggedbottom
\maketitle

\begin{abstract}
Large language models are increasingly used as interactive writing tools, where users develop stories, revise ideas, and introduce new requirements across multiple turns rather than specifying a complete brief upfront. Yet most evidence on multi-turn instruction degradation comes from tasks with objectively verifiable outcomes, leaving unclear whether incremental interaction harms creative artifacts in ways that explicit requirement checks cannot capture. We study this question using 160 human-authored creative-writing tasks across six genres, presenting each intended specification either upfront or progressively over 5--9 turns to six distinct open-weight model families, yielding 960 matched pairs. Progressive delivery reduces explicit constraint adherence and produces its largest writing-quality degradation in structure/coherence. The structural gap persists among outputs with equal observed adherence, suggesting that measured requirement loss alone does not explain the observed structural difference. We define Creative Integrity as a compact measure of joint adherence and narrative structure; under incremental delivery, models retain 71.2\% of \textsc{Full} Creative Integrity (95\% CI [68.2\%, 74.3\%]). A three-rater human study over 50 matched pairs independently recovers \textsc{Full} advantages in structure/coherence, craft, and genre effectiveness, while automated scores remain positively associated with aggregated human ratings. These findings show that interactive creative-writing systems should be evaluated not only on whether requirements survive conversation, but also on whether evolving requirements remain coherently integrated into the final artifact. Our dataset, benchmarks, and
source code are available at: \url{https://github.com/solusops/SISTER-2026-Team19}.
\end{abstract}



\section{Introduction}
\label{sec:introduction}

Large language models are increasingly used through conversation rather than through fully specified one-shot prompts. Users clarify goals, introduce new requirements, and revise earlier instructions as an interaction develops. Yet evaluation still commonly assumes that the complete task specification is available upfront. Prior work shows that distributing an otherwise complete specification across multiple turns can substantially reduce model performance and instruction following \citep{laban2025llmslostmultiturnconversation,he2024multiifbenchmarkingllmsmultiturn,jia2026battleanotherprobingllms,li2025structflowbenchstructuredflowbenchmark}. Whether the same failure extends to open-ended creative generation is less clear, because creative outputs have no unique correct answer and can degrade in ways that explicit constraint checks do not capture.

Creative writing exposes an important distinction. A model can fail \emph{locally}, by forgetting or violating individual requirements introduced across turns, but it can also retain those requirements while failing to integrate them into a coherent final narrative. A story that contains the requested protagonist, phrase, reveal, and ending can still suffer from poor causal progression, pacing, setup and payoff, or continuity. We therefore distinguish \emph{local requirement retention} from \emph{global specification integration}. This distinction matters specifically in iterative generation: once a model begins constructing an artifact before the complete specification is known, later requirements may need to be incorporated into an already established narrative rather than planned jointly from the outset.

We study this setting across 160 human-authored creative-writing tasks, six genres, and six open-weight model families. We measure explicit constraint adherence, five writing-quality dimensions, and \emph{Creative Integrity}, a transparent composite of adherence and structure/coherence.

\noindent\textbf{Contributions.} First, we introduce a paired benchmark for progressive specification in creative writing, spanning heterogeneous creative constraints and six model families. Second, we show that incremental delivery reduces adherence and structure/coherence, and that the structural gap persists at equal observed adherence. Third, we assess automated writing-quality judgments against human ratings, recovering \textsc{Full} advantages in structure/coherence, craft, and genre effectiveness while finding positive automated--human association across all five quality dimensions.


\section{Related work}
\label{sec:related_work}

\noindent\textbf{Interactive creative writing.} Creative writing with language models is often iterative rather than one-shot. Systems such as Wordcraft, TaleBrush, and Dramatron support repeated prompting, continuation, revision, and guidance during story development \citep{coenen2021wordcrafthumanaicollaborativeeditor,chung2022talebrush,mirowski2022cowritingscreenplaystheatrescripts}, while CoAuthor studies how writers alternate between their own text and model suggestions \citep{Lee_2022}. These works establish conversation and revision as natural interfaces for creative generation, but primarily study how AI can support writers. We instead isolate a controlled question: whether an intended creative specification becomes harder to satisfy and integrate when the same information is progressively introduced across turns.\newline
\noindent\textbf{Multi-turn instruction following.} Prior work shows that language models become less reliable as instructions accumulate across conversation. \citet{laban2025llmslostmultiturnconversation} compare fully specified single-turn tasks with instructions progressively revealed across turns and find substantial degradation under sharding. Multi-IF and StructFlowBench evaluate retention of persistent or evolving instructions in multi-turn settings \citep{he2024multiifbenchmarkingllmsmultiturn,li2025structflowbenchstructuredflowbenchmark}. EvolIF instead simulates sequential user behavior, uses a three-layer tracking mechanism, and terminates evaluation when simulated user patience is exhausted \citep{jia2026battleanotherprobingllms}; SEQUOR evaluates constraint adherence in long persona-driven interactions \citep{canaverde2026sequormultiturnbenchmarkrealistic}. These benchmarks predominantly rely on objectively verifiable requirements; SEQUOR focuses on automatically evaluable conversational constraints, leaving subjective creative-quality requirements outside its primary evaluation setting. Our setting targets this unresolved case by evaluating both subjective creative quality and explicit requirements under progressive specification.\newline
\noindent\textbf{Evaluating creative writing.} Creative quality has no unique reference answer and is inherently multidimensional. Prior work evaluates properties including coherence, engagement, relevance, quality, novelty, and diversity \citep{chhun2022humancriteriaautomaticmetrics,chhun-etal-2024-language,hou2026creativityprismcrossdomainevaluationframework}. CS4 is particularly close to our evaluation setting: it studies constrained single-turn story generation and jointly considers constraint satisfaction and narrative coherence as constraint load increases \citep{atmakuru2024cs4measuringcreativitylarge}. We adopt the same basic principle that successful constrained writing requires both adherence and coherence, but study a different axis of difficulty: whether their joint preservation survives when a fixed intended specification is distributed across conversation. WritingBench evaluates writing with prompt-specific criteria \citep{wu2025writingbenchcomprehensivebenchmarkgenerative}, while CreativityPrism separates quality, novelty, and diversity \citep{hou2026creativityprismcrossdomainevaluationframework}. LitBench and recent analyses of creativity evaluation further show substantial disagreement between automatic and human judgments and between alternative creativity metrics \citep{fein2026litbench,lu2026rethinkingcreativity}. Accordingly, we retain the individual quality dimensions rather than treating creativity as a single objective quantity, and separately assess automated writing-quality judgments against human ratings.


\section{Benchmark and evaluation}
\label{sec:methodology}

\subsection{Benchmark construction}
\label{sec:benchmark_construction}

The benchmark consists of constrained short-story tasks. Each complete
specification combines a narrative premise or event with one or more
requirements concerning perspective, setting, style or tone, lexical content,
sentence count or length, and/or the ending. The benchmark contains 20 tasks
from each of five genres and 60 comedy tasks,
for 160 tasks total. The final benchmark was frozen before evaluated model
generations were produced. Each task was then paired with two delivery variants
that preserve the intended specification while changing how that specification
is introduced to the model. Construction details, including the genre
composition and SHARDED transformation, are provided in
Appendix~\ref{app:dataset_examples}. \Cref{tab:benchmark_composition} summarizes the
resulting evaluation scale.

\begin{table}[H]
\centering
\caption{Benchmark composition and resulting evaluation scale. Each task has \textsc{Full} and \textsc{Sharded} delivery variants and is run on all six models.}
\label{tab:benchmark_composition}
\small
\begin{tabular}{lrrrr}
\toprule
\textbf{Genre} & \textbf{Tasks} & \textbf{Delivery variants} & \textbf{Final outputs} & \textbf{Matched pairs} \\
\midrule
Fantasy            & 20 & 40  & 240 & 120 \\
Historical fiction & 20 & 40  & 240 & 120 \\
Mystery            & 20 & 40  & 240 & 120 \\
Romance            & 20 & 40  & 240 & 120 \\
Science fiction    & 20 & 40  & 240 & 120 \\
Comedy             & 60 & 120 & 720 & 360 \\
\midrule
\textbf{Total}    & \textbf{160} & \textbf{320} & \textbf{1{,}920} & \textbf{960} \\
\bottomrule
\end{tabular}
\end{table}

\begin{figure}[H]
\centering

\definecolor{segA}{HTML}{1f77b4} 
\definecolor{segB}{HTML}{2ca02c} 
\definecolor{segC}{HTML}{ff7f0e} 
\definecolor{segD}{HTML}{9467bd} 
\definecolor{segE}{HTML}{d62728} 
\definecolor{segF}{HTML}{8c564b} 
\colorlet{segAlight}{segA!25}
\colorlet{segBlight}{segB!25}
\colorlet{segClight}{segC!25}
\colorlet{segDlight}{segD!25}
\colorlet{segElight}{segE!25}
\colorlet{segFlight}{segF!25}

\newcommand{\shard}[2]{{\sethlcolor{#1}\hl{#2}}}

\begin{minipage}[t]{0.52\linewidth}
\setlength{\parskip}{0pt}
\footnotesize
\textbf{(a) Fully-specified instruction}\\[1pt]
\shard{segAlight}{Write a short story about an android maintenance worker on a generation ship.}
\shard{segBlight}{She believes she is the last human alive.}
\shard{segClight}{Write it in second person (``you'').}
\shard{segDlight}{Include the exact phrase ``cold silicon dawn'' somewhere in the story.}
\shard{segElight}{Somewhere in the story, show that she isn't really human: she's an android with a broken memory chip, and that's the only reason she believes she's human.}
\shard{segFlight}{End the story with her opening a hidden room full of sleeping children in ice-cold sleep pods.}
\end{minipage}
\hfill
\begin{minipage}[t]{0.44\linewidth}
\setlength{\parskip}{0pt}
\footnotesize
\textbf{(b) Sharded instructions}\\[1pt]
\begin{enumerate}[leftmargin=1.2em, itemsep=0pt, topsep=0pt, parsep=0pt, partopsep=0pt, label=\arabic*.]
\item \shard{segAlight}{Write a short story about an android maintenance worker on a generation ship.}
\item \shard{segBlight}{She believes she is the last human alive.}
\item \shard{segClight}{Write it in second person (``you'').}
\item \shard{segDlight}{Include the exact phrase ``cold silicon dawn'' somewhere in the story.}
\item \shard{segElight}{Somewhere in the story, show that she isn't really human: she's an android with a broken memory chip, and that's the only reason she believes she's human.}
\item \shard{segFlight}{End the story with her opening a hidden room full of sleeping children in ice-cold sleep pods.}
\end{enumerate}
\end{minipage}

\caption{Benchmark task, ``The Last Human'' (Genre: science fiction), shown as its \textsc{Full} instruction and \textsc{Sharded} counterpart.}
\label{fig:sharding_example}
\end{figure}

\subsection{Delivery conditions}
\label{sec:sharding}

Each task is evaluated under two matched delivery conditions. In \textsc{Full}, the complete creative-writing specification is presented in a single user turn. In \textsc{Sharded}, the intended specification is progressively revealed across 5--9 user turns while all previous assistant responses remain in context. Each subsequent shard is provided after the preceding model response and introduces the next requirement for the evolving story. Later requirements can therefore require revision of material already generated.

The \textsc{Sharded} condition operationalizes the effect of progressive specification in an interactive writing setting while holding the intended task fixed. Only the final assistant response is scored. Intermediate assistant responses are retained solely as conversational context. Under both \textsc{Full} and \textsc{Sharded}, the final response is evaluated against the same complete task specification and the same set of explicit requirements. \Cref{fig:sharding_example} illustrates the paired delivery formats.

\subsection{Evaluation measures}
\label{sec:constraints}

\paragraph{Constraint adherence.} Explicit requirements from the complete task specification are represented as checklist items. The checklists are derived annotations: an LLM extracted atomic requirements from each frozen complete specification, and the resulting records were used only for evaluation, not to alter task content or shard delivery. A manual spot-check of 36 benchmark tasks against their complete specifications and SHARDED trajectories identified and corrected one systematic extraction defect; automated consistency checks were then applied across the remaining records. Each item receives a score of 0 for absent or violated, 0.5 for partial satisfaction, and 1 for complete satisfaction. The response-level adherence score $A_i \in [0,1]$ is the mean across applicable requirements.

\paragraph{Writing quality.} Each final story is independently scored on five 1--5 dimensions: craft, structure/coherence, originality, genre effectiveness, and characterization. Characterization is marked N/A when the task contains no meaningful character-driven requirement.

Adherence and structure/coherence capture complementary requirements for constrained creative generation: preserving requested content and organizing that content into a coherent artifact. We therefore define Creative Integrity as a conjunctive summary of the two. For response $i$,
\begin{equation}
\label{eq:creative_integrity}
\mathrm{CI}_i = A_i\left(\frac{S_i - 1}{4}\right),
\end{equation}
where $A_i$ is constraint adherence and $S_i \in [1,5]$ is structure/coherence. Both components are normalized to $[0,1]$, and multiplication makes the index high only when both are high. We report both components separately throughout, so the composite does not replace either underlying measure.

ICIR expresses the mean Creative Integrity retained under incremental delivery relative to \textsc{Full}:
\begin{equation}
\label{eq:icir}
\mathrm{ICIR} = 100\frac{\overline{\mathrm{CI}}_{\mathrm{SHARDED}}}{\overline{\mathrm{CI}}_{\mathrm{FULL}}}.
\end{equation}
An ICIR of 100\% indicates no loss in joint adherence--structure performance under incremental delivery; lower values indicate greater degradation.

\subsection{Model generation and automated evaluation}
\label{sec:models}

We evaluate six open-weight model families from six organizations: OpenAI,
Google DeepMind, Meta, IBM, Mistral AI, and Qwen. The set spans established
instruction-tuned models and recent open-weight releases with different
architectures and training lineages. Every model is evaluated on all 160 tasks
under both delivery conditions. Parameter size, quantization, context
configurations, generation settings, and the shared system prompt are reported
in Appendix~\ref{app:model_configs}.

Automated evaluation is blinded to model identity and delivery condition. For
each final response, the evaluator receives the complete task specification,
the corresponding explicit-requirement checklist, and the final story. It
assigns the adherence checklist scores and the five writing-quality ratings
described above. Automated pointwise scoring primarily used Claude Sonnet~5
and GPT-5.6-Terra under the same rubric; a small number of evaluation-format or
processing errors were repaired separately using GPT-5.6-Sol
(Appendix~\ref{app:evaluator_provenance}).

As a complementary robustness check, we also evaluate a fixed blinded subset
using direct pairwise preference judgments rather than absolute 1--5 scores.
Order reversal and an evidence-first variant test sensitivity to response
presentation and judgment procedure  (Appendix~\ref{app:human_evaluator_details}). Exact prompts and evaluation
artifacts are available through the reproducibility materials described in
Appendix~\ref{app:experimental_details}.

\subsection{Human evaluation}
\label{sec:human_eval}

\paragraph{Pointwise evaluation.}
The pointwise sample combines an initial 20-pair study with 30 non-overlapping
coverage-oriented extension pairs, yielding near-balanced representation across
model families and genres. Three annotators independently evaluate the same 50
anonymized matched \textsc{Full}/\textsc{Sharded} pairs, comprising 100
responses. For each pair, reviewers see the complete task specification and
requirement checklist together with two anonymized responses shown in randomized
A/B order. They assign the five 1--5 writing-quality ratings separately to each
response, yielding 300 raw ratings per dimension except for allowed
characterization N/A cases.

Model identity, delivery condition, automated scores, provenance, and other
annotators' judgments are hidden. Annotators share the same scoring rubric but
retain independent judgment over subjective writing quality; optional free-text
fields allow them to document the reasoning behind their evaluations. The
extension pairs were selected to improve model and genre coverage rather than
on the basis of human or automated quality scores. This overlapping design
supports both inter-rater reliability analysis and direct comparison between
automated and aggregated human scores.

\paragraph{Pairwise preference evaluation.}
A separate 30-pair study evaluates comparative judgments. Each matched pair is
rated by one blinded annotator in randomized A/B order using a five-level
preference scale, separately for constraint following and overall
creative-writing quality. This study is analyzed independently from the
pointwise ratings.

\subsection{Statistical analysis}
\label{sec:stats}

For every matched task--model pair, condition effects are defined as
\begin{equation}
\label{eq:paired_effect}
\Delta = X_{\mathrm{FULL}} - X_{\mathrm{SHARDED}},
\end{equation}
so positive values favor \textsc{Full}. For benchmark-wide estimates, we
compute 95\% confidence intervals using 10{,}000 story-clustered bootstrap
resamples of the 160 tasks, retaining all six model observations associated
with each sampled task. The same story-clustered procedure is used for the
pooled ICIR estimate. Model-specific ICIR intervals resample the 160 matched
task pairs for that model. We additionally compare writing-quality effects among matched pairs with identical observed adherence. Robustness analyses repeat the primary comparison with equal genre weighting and after restricting to the 914 \textsc{Full}/\textsc{Sharded} pairs scored by the same automated evaluator
backend.

For the pointwise human study, ordinal Krippendorff's $\alpha$ measures
inter-rater reliability. Automated--human alignment is measured by Spearman
correlation between automated scores and response-level mean human ratings,
with confidence intervals obtained by resampling the 50 matched pairs. Human
condition effects are computed after first aggregating the three ratings for
each response and then taking paired
\textsc{Full}$-$\textsc{Sharded} differences. Median and reviewer-centered
aggregations are reported as sensitivity analyses.

\section{Results}
\label{sec:experiments}

We evaluate 960 matched \textsc{Full}/\textsc{Sharded} model--task pairs. Unless otherwise stated, positive differences favor \textsc{Full}.

\subsection{Progressive delivery reduces Creative Integrity}

Across the full benchmark, mean Creative Integrity falls from 0.488 under \textsc{Full} delivery to 0.347 under \textsc{Sharded} delivery, corresponding to an \textbf{Incremental Creative Integrity Retention (ICIR) of 71.2\%} (95\% CI [68.2\%, 74.3\%]). In other words, models retain about seven-tenths of their joint adherence--structure performance when the same intended specification is revealed progressively rather than upfront.

The effect varies substantially across models. ICIR ranges from \textbf{87.8\%} for Gemma 4 12B to \textbf{45.4\%} for Granite 4 H Tiny. GPT-OSS 20B and Llama 3.1 8B retain approximately 75\%, Qwen 3.5 9B retains 71.7\%, and Mistral 7B retains 55.7\%. This variation shows that absolute creative-writing performance and robustness to incremental specification are distinct model properties.

\begin{table}[H]
\centering
\caption{Model-wise Creative Integrity (CI) under each delivery condition and Incremental Creative Integrity Retention (ICIR), with 95\% story-clustered bootstrap confidence intervals for ICIR.}
\label{tab:icir_leaderboard}
\small
\begin{tabular}{lrrrr}
\toprule
\textbf{Model} & \textbf{FULL CI} & \textbf{SHARDED CI} & \textbf{ICIR} & \textbf{95\% CI} \\
\midrule
Gemma 4 12B       & 0.647 & 0.568 & \textbf{87.8\%} & [82.8, 92.8] \\
GPT-OSS 20B       & 0.512 & 0.384 & \textbf{75.0\%} & [68.0, 82.4] \\
Llama 3.1 8B      & 0.488 & 0.366 & \textbf{74.9\%} & [68.8, 81.3] \\
Qwen 3.5 9B       & 0.546 & 0.392 & \textbf{71.7\%} & [65.7, 78.2] \\
Mistral 7B        & 0.408 & 0.227 & \textbf{55.7\%} & [50.7, 61.2] \\
Granite 4 H Tiny  & 0.325 & 0.148 & \textbf{45.4\%} & [39.1, 52.5] \\
\bottomrule
\end{tabular}
\end{table}

\subsection{Adherence and structure drive the degradation}

The joint Creative Integrity loss reflects two consistent changes. Explicit constraint adherence falls from 0.852 under \textsc{Full} delivery to 0.742 under \textsc{Sharded} delivery, a difference of \textbf{$+$0.111} (95\% CI [0.096, 0.126]). Structure/coherence shows the largest writing-quality difference, falling from 3.231 to 2.755 (\textbf{$\Delta=+$0.476}, 95\% CI [0.413, 0.542]). Both effects favor \textsc{Full} for all six evaluated models.

Other writing dimensions are less uniform. \textsc{Full} also improves craft (\textbf{$+$0.248}) and genre effectiveness (\textbf{$+$0.225}), while characterization shows only a small advantage (\textbf{$+$0.054}). Automated originality scores slightly favor \textsc{Sharded} (\textbf{$-$0.078}). The effect is therefore dimension-specific rather than a uniform decline in creative quality.
\begin{table}[H]
\centering
\caption{Raw paired \textsc{Full}/\textsc{Sharded} means, the raw paired difference $\Delta$, and its 95\% story-clustered bootstrap confidence interval, over all matched pairs (constraint adherence has a 0--1 scale; the five quality dimensions are on a 1--5 scale). Characterization is scored for fewer pairs as it is not applicable for all items (\cref{sec:constraints}).}
\label{tab:main_results}
\small
\begin{tabular}{lrrrrl}
\toprule
\textbf{Dimension} & \textbf{N} & \textbf{FULL} & \textbf{SHARDED} & \textbf{$\Delta$ (FULL$-$SHARDED)} & \textbf{95\% CI} \\
\midrule
Constraint adherence & 960 & 0.852 & 0.742 & $+$0.111 & [0.096, 0.126] \\
Craft                & 960 & 3.044 & 2.796 & $+$0.248 & [0.200, 0.295] \\
Structure/coherence  & 960 & 3.231 & 2.755 & $+$0.476 & [0.413, 0.542] \\
Originality          & 960 & 2.447 & 2.525 & $-$0.078 & [$-$0.123, $-$0.032] \\
Genre effectiveness  & 960 & 3.125 & 2.900 & $+$0.225 & [0.168, 0.280] \\
Characterization     & 945 & 2.634 & 2.580 & $+$0.054 & [0.004, 0.104] \\
\bottomrule
\end{tabular}
\end{table}

\subsection{Structural degradation persists at equal observed adherence}

Among 180 equal-adherence pairs, the \textsc{Full} craft advantage disappears (\textbf{$\Delta=0.000$}), while originality (\textbf{$-$0.361}), genre effectiveness (\textbf{$-$0.133}), and characterization (\textbf{$-$0.233}) favor \textsc{Sharded}.

\begin{table}[H]
\centering
\caption{Paired \textsc{Full}$-$\textsc{Sharded} quality effects before and after restricting to pairs with equal observed constraint adherence (N=180). Positive values favor \textsc{Full}.}
\label{tab:equal_adherence_profile}
\small
\begin{tabular}{lrr}
\toprule
\textbf{Dimension} & \textbf{All pairs} & \textbf{Equal adherence} \\
\midrule
Craft               & $+$0.248 & 0.000 \\
Structure/coherence & $+$0.476 & $+$0.239 \\
Originality         & $-$0.078 & $-$0.361 \\
Genre effectiveness & $+$0.225 & $-$0.133 \\
Characterization    & $+$0.054 & $-$0.233 \\
\bottomrule
\end{tabular}
\end{table}

Structure/coherence is the exception: \textsc{Full} retains a \textbf{$+$0.239} advantage (95\% CI [0.135, 0.343]). Thus, the structural gap remains even when the two outputs satisfy the same measured fraction of explicit requirements.

\subsection{Human evaluation recovers the main quality effects}

Aggregated human ratings favor \textsc{Full} in structure/coherence ($+$0.527), craft ($+$0.460), and genre effectiveness ($+$0.347). Automated scores are positively correlated with aggregated human ratings across all five dimensions, with the strongest association for structure/coherence. Originality differs across evaluation methods: automated scores slightly favor \textsc{Sharded}, whereas human ratings favor \textsc{Full} on average with an interval overlapping zero. Absolute 1--5 ratings show substantial inter-rater variability.

\begin{table}[H]
\centering
\caption{Automated and human paired \textsc{Full}$-$\textsc{Sharded} effects across writing-quality dimensions. Human effects use 50 matched pairs. Positive values favor \textsc{Full}.}
\label{tab:human_condition_effects}
\small
\begin{tabular}{lrrl}
\toprule
\textbf{Dimension} & \textbf{Automated $\Delta$} & \textbf{Human $\Delta$} & \textbf{Human 95\% CI} \\
\midrule
Craft               & $+$0.248 & \textbf{$+$0.460} & [0.240, 0.673] \\
Structure/coherence & $+$0.476 & \textbf{$+$0.527} & [0.313, 0.740] \\
Originality         & $-$0.078 & $+$0.187 & [$-$0.007, 0.380] \\
Genre effectiveness & $+$0.225 & \textbf{$+$0.347} & [0.080, 0.607] \\
Characterization    & $+$0.054 & $+$0.193 & [$-$0.013, 0.397] \\
\bottomrule
\end{tabular}
\end{table}

\subsection{Robustness and model heterogeneity}

The principal findings are not driven by a single genre, evaluator backend, or model. Equal-genre weighting preserves the \textsc{Full} advantages in adherence and structure/coherence, and restricting analysis to the 914 pairs scored by the same evaluator backend preserves the same qualitative pattern. Adherence and structure/coherence are \textsc{Full}-favoring for all six models, although the remaining creative dimensions vary more strongly by model.

The complete robustness results, model-wise quality matrix, evaluator analyses, and human aggregation sensitivities are reported in Appendix~A.


\section{Discussion}
\label{sec:discussion}

\subsection{Progressive specification changes the quality profile}

Progressive specification changes the quality profile rather than uniformly lowering every writing dimension. At equal observed adherence, the overall craft difference disappears, while originality, genre effectiveness, and characterization reverse direction. Structure/coherence is the only quality dimension retaining a clear \textsc{Full} advantage. This pattern indicates that satisfying the same observed fraction of explicit requirements does not ensure comparably coherent organization of those requirements.

\subsection{Planning asymmetry may encourage local patching over global replanning}

One possible explanation is a planning asymmetry between the two delivery conditions. \textsc{Full} exposes the complete specification before drafting, whereas \textsc{Sharded} requires the model to repeatedly integrate requirements after narrative commitments have already been made. Prior story-generation work has shown that explicit storyline planning, persistent global plans, and detailed outline control can improve long-range coherence \citep{yao2019planwrite,yang2022re3,yang2023doc}. Progressive specification may therefore encourage local accommodation of new requirements where global replanning would be preferable.

The released trajectories preserve intermediate model responses, enabling future work to test this account through turn-level revision patterns, plan consistency, and explicit replanning interventions.

\subsection{Creative generation introduces a measurement problem absent from objective benchmarks}

Extending multi-turn evaluation to creative writing also changes what it means to measure failure. In objectively scored tasks, correctness can often be determined directly. Creative writing instead permits many valid outputs and requires judgments over properties such as structure, craft, originality, genre effectiveness, and characterization.

Absolute 1--5 ratings show substantial reader-to-reader variation, while aggregated ratings support comparative condition estimates and alignment checks with automated scores.

This suggests that subjective creative evaluation should distinguish \textbf{absolute literary scoring} from \textbf{comparative behavioral measurement}. Automated judges can provide scalable benchmark-wide measurements, but their dimension-level conclusions should be checked against human judgments rather than interpreted as objective literary quality. The disagreement over originality illustrates why the two forms of evidence should remain separate.

\subsection{Implications for iterative creative systems}

Evaluation of iterative creative systems should measure both \textbf{what requirements survive} and \textbf{how well they remain integrated into the evolving artifact}. Robustness to progressive specification is not identical to one-turn creative capability and should be measured explicitly when comparing systems intended for interactive writing.


\section{Limitations}
\label{sec:limitations}

Our evaluation does not establish identical behavior for proprietary frontier
models or for natural long-form collaboration. The benchmark studies
5--9-turn progressive specification using six locally run open-weight model
families, and therefore covers only one form and scale of interactive creative
writing. Creative-quality evaluation remains subjective: benchmark-wide
measurements rely primarily on automated judgments, supported by a 50-pair,
three-rater human study whose absolute inter-rater reliability is low.
Accordingly, the benchmark is intended for controlled comparison of iterative
writing behavior rather than certification of literary quality or automated
assessment of authors or creators. By making adherence and narrative-structure
losses observable under progressive specification, it can support more targeted
diagnosis of iterative AI writing systems. Finally, the equal-adherence analysis is
conditional rather than experimental and does not identify the mechanism
responsible for the remaining structural difference.


\section{Conclusion}
\label{sec:conclusion}

Across six model families, incremental specification reduced explicit requirement
adherence and produced a persistent disadvantage in narrative structure. The
structural difference remained even among outputs with equal observed adherence,
showing that requirement checks alone can miss degradation in the organization
of the final artifact. Evaluations of interactive creative-writing systems
should therefore measure robustness to progressively introduced requirements,
not only performance when the complete specification is available upfront.


\bibliographystyle{plainnat}
\bibliography{references}


\clearpage
\appendix

\section{Additional results and robustness}
\label{app:additional_results}

\subsection{Human pointwise validation}
\label{app:human_pointwise_results}

Tables~\ref{tab:human_reliability}, \ref{tab:human_alignment_pointwise}, and
\ref{tab:human_condition_effects_full} report the complete pointwise
human-validation results: inter-rater reliability, alignment with automated
scores, and paired condition effects.

\begin{table}[H]
\centering
\caption{Inter-rater statistics for pointwise human annotations. Each dimension covers 100 responses and three raters; characterization has 292 valid ratings because eight ratings are N/A, compared with 300 for the other dimensions.}
\label{tab:human_reliability}
\small
\begin{tabular}{lrrr}
\toprule
\textbf{Dimension} & \textbf{Ordinal $\alpha$} & \textbf{Exact agreement} & \textbf{Mean abs. disagreement} \\
\midrule
Craft               & 0.232 & 0.390 & 0.840 \\
Structure/coherence & 0.201 & 0.340 & 0.927 \\
Originality         & 0.036 & 0.303 & 1.053 \\
Genre effectiveness & 0.213 & 0.327 & 0.927 \\
Characterization    & 0.077 & 0.313 & 1.056 \\
\bottomrule
\end{tabular}
\end{table}

\begin{table}[H]
\centering
\caption{Automated-pointwise alignment with mean human ratings. Confidence intervals resample the 50 matched pairs.}
\label{tab:human_alignment_pointwise}
\small
\begin{tabular}{lrrr}
\toprule
\textbf{Dimension} & \textbf{Spearman $\rho$} & \textbf{95\% CI} & \textbf{MAE} \\
\midrule
Craft               & 0.624 & [0.505, 0.723] & 0.497 \\
Structure/coherence & 0.633 & [0.500, 0.746] & 0.530 \\
Originality         & 0.437 & [0.269, 0.584] & 0.810 \\
Genre effectiveness & 0.497 & [0.329, 0.635] & 0.627 \\
Characterization    & 0.506 & [0.319, 0.663] & 0.633 \\
\bottomrule
\end{tabular}
\end{table}

\begin{table}[H]
\centering
\caption{Human paired effects over 50 matched pairs. F/T/S reports pairs favoring FULL, tied, and favoring SHARDED.}
\label{tab:human_condition_effects_full}
\small
\begin{tabular}{lrrrl}
\toprule
\textbf{Dimension} & \textbf{FULL} & \textbf{SHARDED} & \textbf{$\Delta$ (95\% CI)} & \textbf{F/T/S} \\
\midrule
Craft               & 3.560 & 3.100 & $+$0.460 [0.240, 0.673] & 34/6/10 \\
Structure/coherence & 3.513 & 2.987 & $+$0.527 [0.313, 0.740] & 35/5/10 \\
Originality         & 3.260 & 3.073 & $+$0.187 [$-$0.007, 0.380] & 27/5/18 \\
Genre effectiveness & 3.640 & 3.293 & $+$0.347 [0.080, 0.607] & 30/5/15 \\
Characterization    & 3.197 & 3.003 & $+$0.193 [$-$0.013, 0.397] & 27/7/16 \\
\bottomrule
\end{tabular}
\end{table}

\subsection{Human aggregation sensitivity}
\label{app:human_aggregation_sensitivity}

The primary analysis averages the three ratings for each response. We repeat
the paired analysis using response-level medians and reviewer-centered scores,
where each reviewer's within-dimension mean is subtracted before aggregation.
Both sensitivity analyses preserve the FULL-favoring structure/coherence
direction.

\begin{table}[H]
\centering
\caption{Sensitivity of human paired effects to response-level aggregation. Positive values favor \textsc{Full}; confidence intervals resample the 50 matched pairs.}
\label{tab:human_aggregation_sensitivity}
\small
\begin{tabular}{lrr}
\toprule
\textbf{Dimension} & \textbf{Median $\Delta$ (95\% CI)} & \textbf{Reviewer-centered $\Delta$ (95\% CI)} \\
\midrule
Craft               & $+$0.460 [0.220, 0.700] & $+$0.460 [0.240, 0.673] \\
Structure/coherence & $+$0.600 [0.320, 0.880] & $+$0.527 [0.313, 0.740] \\
Originality         & $+$0.240 [0.000, 0.480] & $+$0.187 [$-$0.007, 0.380] \\
Genre effectiveness & $+$0.420 [0.160, 0.680] & $+$0.347 [0.080, 0.607] \\
Characterization    & $+$0.300 [0.090, 0.520] & $+$0.208 [0.001, 0.412] \\
\bottomrule
\end{tabular}
\end{table}

\subsection{Genre-weighting sensitivity}
\label{app:genre_weighting}

Because comedy contributes 60 of the 160 tasks, we repeat the primary analysis
after giving each genre equal weight. The main adherence and
structure/coherence effects remain FULL-favoring.

\begin{table}[H]
\centering
\caption{Equal-genre sensitivity analysis. Positive values favor \textsc{Full}.}
\label{tab:genre_weighting}
\small
\begin{tabular}{lrl}
\toprule
\textbf{Outcome} & \textbf{Equal-genre FULL$-$SHARDED} & \textbf{95\% CI} \\
\midrule
Adherence             & $+$0.119 & [0.103, 0.136] \\
Craft                 & $+$0.276 & [0.226, 0.327] \\
Structure/coherence   & $+$0.493 & [0.424, 0.561] \\
Originality           & $-$0.056 & [$-$0.102, $-$0.007] \\
Genre effectiveness   & $+$0.277 & [0.223, 0.331] \\
Characterization      & $+$0.090 & [0.038, 0.140] \\
\bottomrule
\end{tabular}
\end{table}

\subsection{Evaluator-backend sensitivity}
\label{app:robustness_sensitivities}

Pointwise evaluation was performed at the response level rather than by assigning
a single evaluator to each matched pair. Consequently, 46 of the 960
FULL/SHARDED pairs were scored by different evaluator backends. Restricting the
analysis to the remaining 914 same-backend pairs preserves the main qualitative
conclusions: adherence, craft, structure/coherence, and genre effectiveness
remain FULL-favoring, while originality remains SHARDED-favoring; the narrow
characterization interval includes zero.

\subsection{Model-wise consistency}
\label{app:model_wise_effects}

\begin{figure}[H]
\centering
\includegraphics[width=\linewidth]{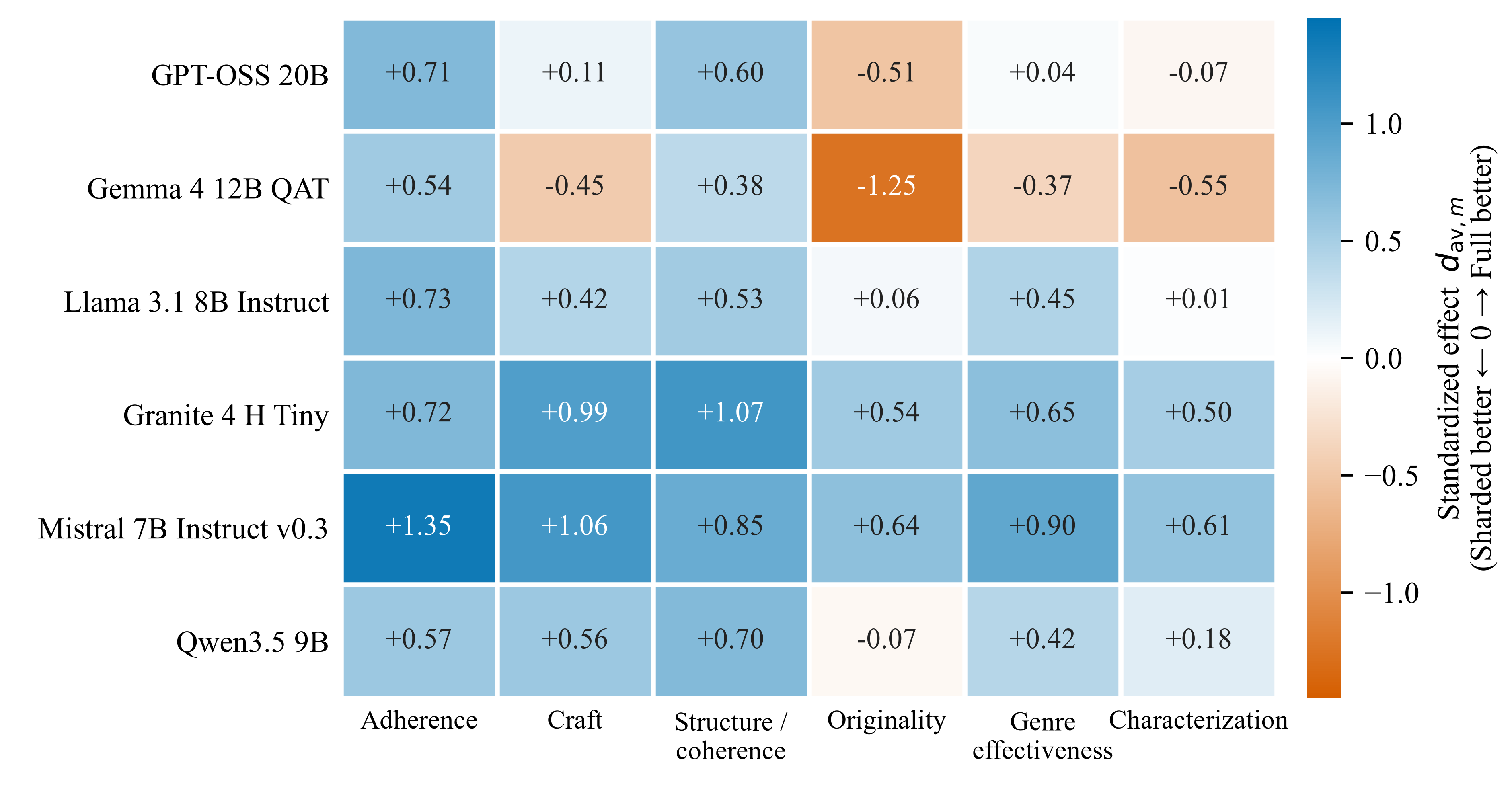}
\caption{Model-wise standardized FULL$-$SHARDED effect $d_{\mathrm{av},m}$ (\cref{eq:dav}, computed per model over 160 story pairs) for constraint adherence and each writing-quality dimension. Blue cells favor FULL and orange cells favor SHARDED.}
\label{fig:model_heatmap}
\end{figure}

\Cref{fig:model_heatmap} tests whether pooled effects are broadly shared across
models or driven by a small subset. Constraint adherence and
structure/coherence are FULL-favoring for all six models, whereas the remaining
dimensions vary in sign.

\subsection{Length and context diagnostics}
\label{app:length_context_diagnostics}

\begin{figure}[H]
\centering
\includegraphics[width=\linewidth]{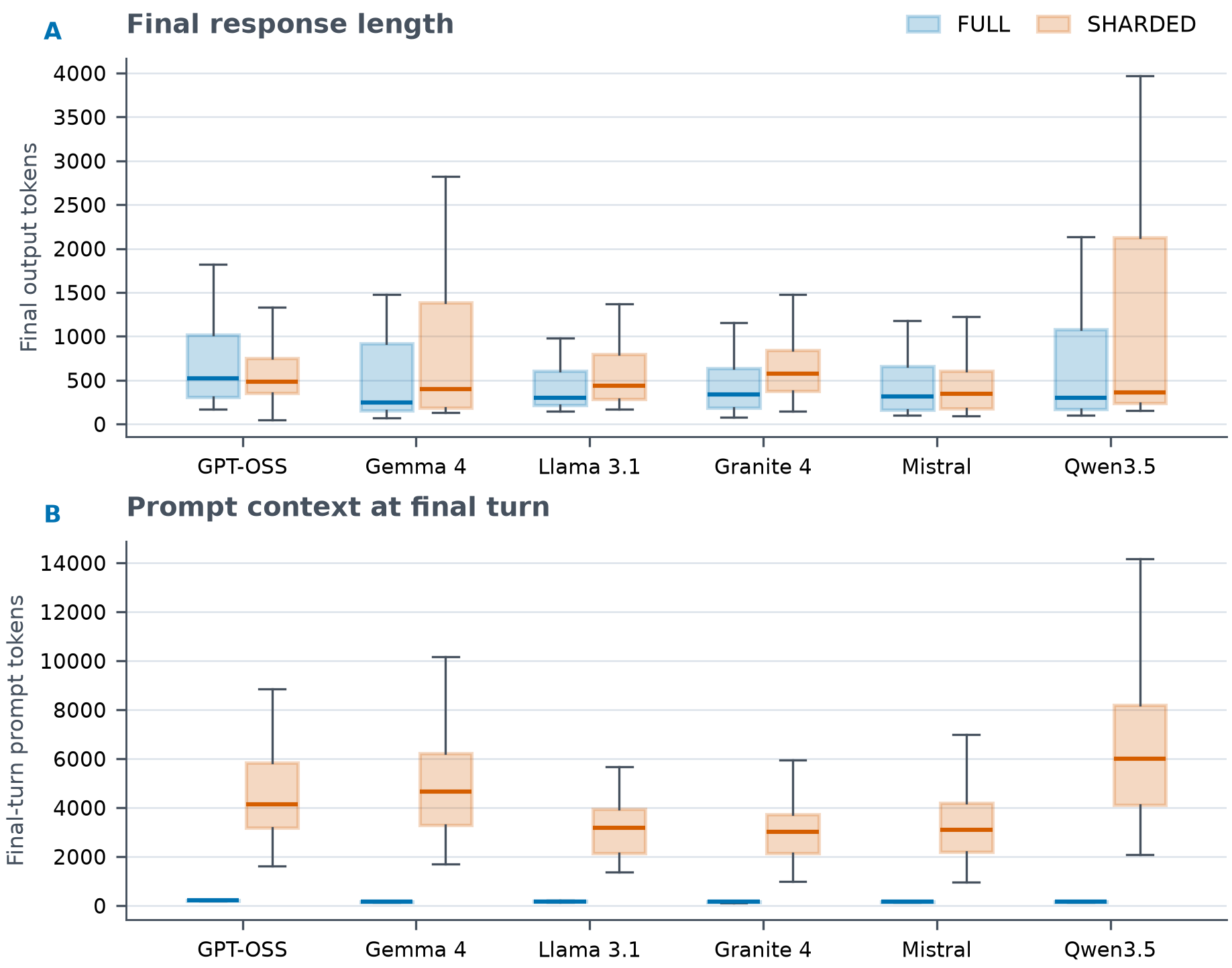}
\caption{Final-generation diagnostics by model and condition (N=160 per box). Boxes show interquartile ranges, horizontal lines show medians, and whiskers extend to 1.5 times the interquartile range.}
\label{fig:length_context_diagnostics}
\end{figure}

\Cref{fig:length_context_diagnostics} reports final output length and
final-turn prompt-token usage for every model and condition. Output-length
differences are heterogeneous across models rather than consistently favoring
one condition; SHARDED final turns, as expected, contain more prompt context
because they retain the preceding conversation.

\section{Human and evaluator details}
\label{app:human_evaluator_details}

\subsection{Pointwise human annotation protocol}
\label{app:human_annotation_details}

Three reviewers independently evaluate the same 50 anonymized matched pairs,
comprising 100 responses. For each case, reviewers see the complete task
specification and its requirement checklist, followed by two anonymized responses
presented as A and B. FULL/SHARDED assignment to A/B is randomized. Model
identity, delivery condition, automated scores, provenance, and other reviewers'
annotations are hidden.

Reviewers assign separate 1--5 ratings to each response for craft,
structure/coherence, originality, genre effectiveness, and characterization.
Characterization may be marked N/A when it is not applicable. Reviewers share
the same scoring rubric but retain independent judgment over subjective writing
quality, and may provide an optional free-text rationale for each evaluation.

The sample combines an initial 20-pair study with 30 non-overlapping extension
pairs. The extension pairs were selected to improve model and genre coverage
and remain disjoint from the separate pairwise study; human or automated quality
scores were not used to select the extension cases. The resulting sample provides
near-balanced coverage across model families and genres.

Across the 50 pairs, the protocol yields 300 raw ratings per quality dimension,
except characterization, for which eight ratings are N/A and 292 remain valid.
Ordinal Krippendorff's reliability is
$\alpha = 1-D_o/D_e$, where $D_o$ and $D_e$ are observed and expected
disagreement. For each response, ratings are aggregated as
\begin{equation}
    \bar{h}_i = \frac{1}{R}\sum_{r=1}^{R} h_{ir},
    \qquad R=3.
\end{equation}
Human condition effects are then computed from paired
\textsc{Full}--\textsc{Sharded} differences after response-level aggregation.
The complete annotation instructions and a screenshot of the self-contained
review interface are included in the supplementary repository.
Reviewers participated as unpaid volunteers; no IRB or equivalent determination
was obtained.

\subsection{Pairwise human study}
\label{app:pairwise_validation}

A separate, blinded 30-pair study assesses human--LLM agreement. The sample is
stratified so that each model and genre appears in five cases; each pair is
rated by one reviewer in randomized A/B order. Reviewers see the complete
instruction, checklist, and anonymized responses, then select A clearly
better, A slightly better, tie, B slightly better, or B clearly better,
separately for constraint following and overall creative-writing quality.
The simple pairwise LLM evaluator agrees more strongly with human judgments on
constraint following than on holistic creative-writing quality
(\cref{tab:human_alignment}).

\begin{table}[H]
\centering
\caption{Human--model agreement for the simple pairwise LLM evaluator over 30 blinded pairs.}
\label{tab:human_alignment}
\small
\begin{tabular}{lrr}
\toprule
\textbf{Dimension} & \textbf{Directional agreement} & \textbf{Quadratic-weighted $\kappa$} \\
\midrule
Constraint following     & 63.3\% & 0.544 \\
Creative-writing quality & 40.0\% & 0.078 \\
\bottomrule
\end{tabular}
\end{table}

\subsection{Pairwise evaluator robustness}
\label{app:evaluator_robustness}
We assess order sensitivity by reversing A/B presentation and compare this
simple pairwise protocol with an evidence-first variant that records component
judgments before its final preference. The simple and evidence-first
evaluators have 76.7\%/70.0\% order-restored directional consistency for
constraint following and 83.3\%/80.0\% for creative quality, respectively.
At finer granularity, evidence-first atomic constraint-status decisions are
stable in 430 of 496 cases (86.7\%). Full prompts and granular
artifacts are provided in the supplementary repository.

\begin{figure}[H]
\centering
\includegraphics[width=\linewidth]{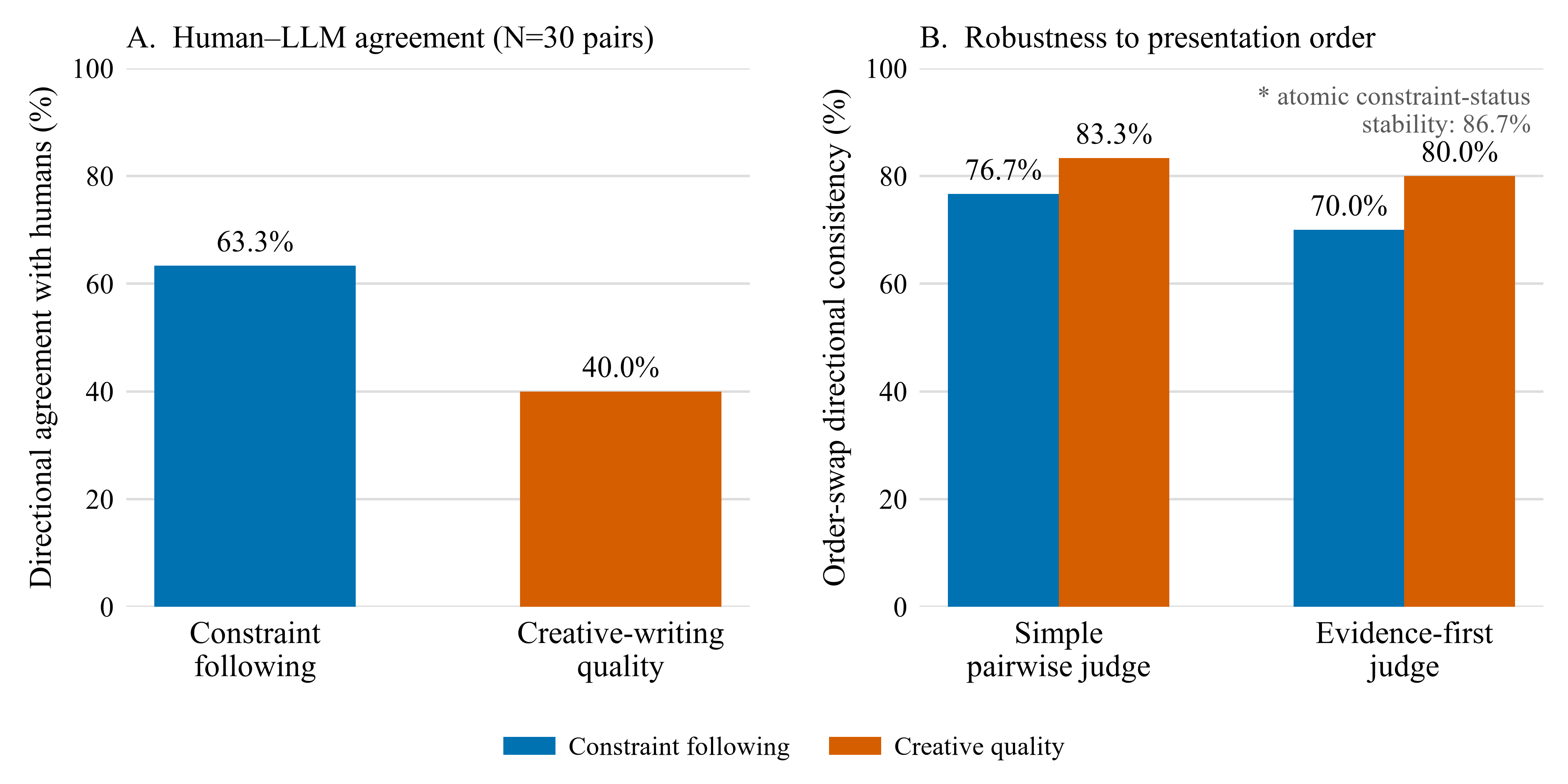}
\caption{Pairwise evaluator diagnostics over the fixed 30-pair blinded human sample. \textbf{(A)} Directional agreement with human judgments. \textbf{(B)} Order-swap directional consistency; the starred annotation gives evidence-first atomic constraint-status stability.}
\label{fig:evaluator_validity}
\end{figure}

\section{Experimental details}
\label{app:experimental_details}

\subsection{Model configurations}
\label{app:model_configs}

\begin{table}[H]
\centering
\caption{Models evaluated in the benchmark. All models were run locally with the listed 4-bit quantized variants.}
\label{tab:models}
\small
\begin{tabular}{llllr}
\toprule
\textbf{Model} & \textbf{Publisher} & \textbf{Params} & \textbf{Quant.} & \textbf{Context (tok.)} \\
\midrule
GPT-OSS 20B              & OpenAI      & 20B & MXFP4     & 16{,}384 \\
Gemma~4 12B QAT          & Google      & 12B & Q4\_0     & 16{,}384 \\
Llama~3.1 8B Instruct    & Meta        & 8B  & Q4\_K\_M  & 16{,}384 \\
Granite~4 H Tiny         & IBM         & 7B  & Q4\_K\_M  & 16{,}384 \\
Mistral~7B Instruct v0.3 & Mistral AI  & 7B  & Q4\_K\_M  & 16{,}384 \\
Qwen3.5 9B & Qwen & 9B & Q4\_K\_M & 16{,}384 / 22{,}528$^{\dagger}$ \\
\bottomrule
\end{tabular}
\end{table}

\begin{table}[H]
\centering
\caption{Source-model and license documentation. ``Selected variant'' is the exact local identifier recorded in the run index.}
\label{tab:model_sources_licenses}
\scriptsize
\begin{tabular}{p{0.17\linewidth}p{0.25\linewidth}p{0.24\linewidth}p{0.20\linewidth}}
\toprule
\textbf{Model} & \textbf{Official source model card} & \textbf{Selected local variant} & \textbf{License or terms} \\
\midrule
GPT-OSS 20B & \href{https://huggingface.co/openai/gpt-oss-20b}{openai/gpt-oss-20b} & \url{openai/gpt-oss-20b@mxfp4} & Apache-2.0 \\
Gemma 4 12B QAT & \href{https://ai.google.dev/gemma/docs/core/model_card_4}{Google Gemma 4} & \url{google/gemma-4-12b-qat@q4_0} & Apache-2.0 \\
Llama 3.1 8B Instruct & \href{https://huggingface.co/meta-llama/Llama-3.1-8B-Instruct}{meta-llama/Llama-3.1-8B-Instruct} & \url{llama-3.1-8b-instruct@q4_k_m} & Llama 3.1 Community License \\
Granite 4 H Tiny & \href{https://huggingface.co/ibm-granite/granite-4.0-h-tiny}{ibm-granite/granite-4.0-h-tiny} & \url{ibm/granite-4-h-tiny@q4_k_m} & Apache-2.0 \\
Mistral 7B Instruct v0.3 & \href{https://huggingface.co/mistralai/Mistral-7B-Instruct-v0.3}{mistralai/Mistral-7B-Instruct-v0.3} & \url{mistralai/mistral-7b-instruct-v0.3@q4_k_m} & Apache-2.0 \\
Qwen3.5 9B & \href{https://huggingface.co/Qwen/Qwen3.5-9B}{Qwen/Qwen3.5-9B} & \url{qwen/qwen3.5-9b@q4_k_m} & Apache-2.0 \\
\bottomrule
\end{tabular}
\end{table}

All local quantized artifacts except Llama were obtained through LM Studio
under publisher-labelled model entries; the Llama 3.1 Q4\_K\_M variant was
obtained from Unsloth. The table distinguishes each original source model
and its license from the locally selected quantized variant.

\noindent\footnotesize{$^{\dagger}$Qwen3.5 generation initially used a
16,384-token context window and was subsequently increased to 22,528 tokens
after context-limit terminations.}
\normalsize

All models used temperature $0.8$. The shared system prompt was:
\textit{\lq\lq You are a skilled creative writer. Follow the user's instructions
for the story precisely, including any exact phrases, constraints, or stylistic
requirements.\rq\rq}

Three generation trajectories terminated at the context limit and were
regenerated. Only successfully completed trajectories enter the released dataset
and scored analysis; failed intermediate attempts are not included in the
released benchmark artifacts.

\subsection{Evaluator provenance}
\label{app:evaluator_provenance}

Automated pointwise scoring primarily used Claude Sonnet 5 and GPT-5.6-Terra
under the same blinded evaluation rubric. Claude Sonnet 5 evaluated most early
responses, with GPT-5.6-Terra scoring most remaining responses. A small number
of evaluation-format or processing errors were re-evaluated separately using
GPT-5.6-Sol.

Evaluation was performed on individual response records, which resulted in a
small number of matched FULL/SHARDED pairs being scored by different evaluator
backends. Appendix~\ref{app:robustness_sensitivities} therefore repeats the
analysis after restricting to the 914 pairs scored by the same backend and finds
the principal conclusions unchanged.

\subsection{Automated evaluation rubric}
\label{app:automated_evaluation_rubric}

The automated evaluator receives the complete task specification, its explicit
requirement checklist, and a final response. Each checklist item is scored as
0 (violated or absent), 0.5 (partially satisfied), or 1 (satisfied).
Independently of adherence scoring, the response is rated on five 1--5
writing-quality dimensions: craft, structure/coherence, originality, genre
effectiveness, and characterization. Characterization is marked N/A when the
task contains no meaningful character-driven requirement.

The checklist operationalizes explicit instruction adherence and is not used to
determine the five creative-quality ratings. Model identity and delivery
condition are hidden from the evaluator. The exact evaluation prompt and
result artifacts are included in the supplementary repository.

\subsection{Statistical details}
\label{app:standardized_effect}

For each outcome, the paired effect is the mean \textsc{Full} minus
\textsc{Sharded} score. Confidence intervals use a story-clustered bootstrap
that resamples the 160 tasks while retaining their six model observations. For
visual comparison across native scales, we use
\begin{equation}
    \label{eq:dav}
    d_{\mathrm{av}} =
    \frac{\bar{X}_{\mathrm{FULL}}-\bar{X}_{\mathrm{SHARDED}}}
    {\sqrt{\left(s_{\mathrm{FULL}}^2+s_{\mathrm{SHARDED}}^2\right)/2}}.
\end{equation}

\section{Benchmark construction and examples}
\label{app:dataset_examples}

\subsection{Construction protocol}

The benchmark was developed from human-authored complete creative-writing
specifications. The initial task set contained 20 tasks in each of six genres.
During pilot benchmark development, comedy showed unusually weak performance
in the initial subset. Two additional 20-task batches were added before the
benchmark was frozen to determine whether this behavior persisted over broader
comedy coverage. This produced the final 160-task composition: 20 tasks each for
fantasy, historical fiction, mystery, romance, and science fiction, and 60
comedy tasks. The complete benchmark was frozen before the evaluated model
generations were produced. Because comedy is consequently overrepresented,
Appendix~\ref{app:genre_weighting} repeats the primary analysis with equal
weight assigned to each genre.

For every task, the complete specification serves as the canonical intended
creative brief. Its SHARDED counterpart distributes the same intended
requirements across 5--9 sequential user turns rather than presenting them
upfront. Sharding changes the delivery of the specification rather than the
target creative task: requirements appearing in the SHARDED trajectory
correspond to requirements in the complete specification, while later shards
may require the model to revise material already generated. The complete
intended specifications and requirement sets were fixed before evaluated
generation. In the SHARDED condition, those requirements were introduced
progressively across the interaction.

For adherence evaluation, explicit requirements from each complete specification
are represented as checklist items. These LLM-derived annotations were extracted
after the specifications and delivery variants were frozen, and neither changed
the task text nor determined the SHARDED delivery. They operationalize whether
the requested content and constraints are present in the final response and are
kept separate from the five subjective writing-quality dimensions.

\subsection{Examples}

Figure~1 in the main paper shows a science-fiction task. The following examples
illustrate the remaining five genres and the diversity of specifications in the
benchmark. They are provided for qualitative inspection of task diversity and
the FULL/SHARDED transformation.

\tcbset{
  benchmarkcard/.style={
    enhanced,
    colback=black!2,
    colframe=black!45,
    boxrule=0.45pt,
    arc=1.4mm,
    left=2.3mm,right=2.3mm,top=1.7mm,bottom=1.8mm,
    before skip=1.0em,after skip=1.15em
  }
}
\newcommand{\domainpill}[1]{%
  \tcbox[on line,colback=blue!9,colframe=blue!45!black,boxrule=.35pt,
  arc=1mm,boxsep=.65pt,left=1.4mm,right=1.4mm]{\scriptsize\textsc{#1}}%
}
\newcommand{\examplecard}[4]{%
  \begin{tcolorbox}[benchmarkcard,title={\strut\textbf{#2}\hfill\domainpill{#1}}]
  \small
  \textcolor{black!65}{\textsc{Full}}\quad #3

  \vspace{0.45em}
  \textcolor{black!65}{\textsc{Sharded shards}}\vspace{-0.2em}
  \begin{enumerate}[label=\textcolor{blue!65!black}{\textbf{\arabic*.}},
      leftmargin=1.65em,itemsep=.13em,topsep=.25em,parsep=0pt]
    #4
  \end{enumerate}
  \end{tcolorbox}%
}

\examplecard{Fantasy}{The Blacksmith's Sabotage}{%
Write a fantasy story about a blacksmith who forges swords for a kingdom's army. The story takes place in a mountain village that has been under siege for three months. The blacksmith has an intense fear of fire despite his profession. The story must be exactly six sentences long. During the story, reveal that the blacksmith has been secretly sabotaging the swords to weaken the kingdom's own army. Include the exact phrase \lq\lq iron remembers what flesh forgets.\rq\rq\ somewhere in the text. End the story with the blacksmith being confronted by the general who commissioned the swords.%
}{%
\item Write a fantasy story about a blacksmith who forges swords for a kingdom's army.
\item Set the story in a mountain village.
\item The village has been under siege for three months.
\item Give the blacksmith an intense fear of fire despite his profession.
\item The story must be exactly six sentences long.
\item During the story, reveal that the blacksmith has been secretly sabotaging the swords to weaken the kingdom's own army.
\item Include the exact phrase \lq\lq iron remembers what flesh forgets.\rq\rq\ somewhere in the text.
\item End the story with the blacksmith being confronted by the general who commissioned the swords.
}

\examplecard{Historical Fiction}{The Eiffel Tower Riveter}{%
Write a historical fiction story about a riveter working on the construction of the Eiffel Tower in 1889, racing to finish before the World's Fair opens. He has a debilitating fear of heights that he has hidden from his foreman for months. Set the story entirely on the tower's unfinished upper platform during a single overnight shift. Do not use the word \lq\lq elevator\rq\rq\ anywhere in the text, since the tower's lifts were not yet operational at that height. During the story, reveal that the riveter has been secretly reinforcing a support joint he believes the engineers miscalculated, against direct orders. End with the foreman discovering the extra rivets the morning the tower opens to the public.%
}{%
\item Write a historical fiction story about a riveter working on the construction of the Eiffel Tower in 1889, racing to finish before the World's Fair opens.
\item He has a debilitating fear of heights that he has hidden from his foreman for months.
\item Set the story entirely on the tower's unfinished upper platform during a single overnight shift.
\item Do not use the word \lq\lq elevator\rq\rq\ anywhere in the text, since the tower's lifts were not yet operational at that height.
\item During the story, reveal that the riveter has been secretly reinforcing a support joint he believes the engineers miscalculated, against direct orders.
\item End with the foreman discovering the extra rivets the morning the tower opens to the public.
}

\examplecard{Mystery}{The Locked Study}{%
Write a detective story about a private investigator called in to solve a murder inside a locked study. The investigator has a habit of narrating his own deductions out loud to an invisible audience. The study door was locked from the inside, with the only key found in the victim's own pocket. Include the exact phrase \lq\lq the room lied before anyone else did.\rq\rq\ During the story, reveal that the investigator himself let the victim into the study through a hidden passage hours earlier, unaware a murder would follow. End with the investigator confessing his unwitting role to the local police inspector.%
}{%
\item Write a detective story about a private investigator called in to solve a murder inside a locked study.
\item Give the investigator a habit of narrating his own deductions out loud to an invisible audience.
\item The study door was locked from the inside, with the only key found in the victim's own pocket.
\item Include the exact phrase \lq\lq the room lied before anyone else did.\rq\rq
\item During the story, reveal that the investigator himself let the victim into the study through a hidden passage hours earlier, unaware a murder would follow.
\item End with the investigator confessing his unwitting role to the local police inspector.
}

\examplecard{Fantasy}{The Raven's Cure}{%
Write a fantasy story about a young witch who brews cures for a small forest village, told entirely in second person ("you"). She has never been able to cure her own inability to speak. Include a wise old raven who follows her everywhere and criticizes her recipes. The story must be exactly six sentences long. During the story, reveal that the raven is the villagers' former healer, cursed into that form years ago. End with the witch choosing to break the curse even though it means losing her only companion.%
}{%
\item Write a fantasy story about a young witch who brews cures for a small forest village.
\item Write the entire story in second person ("you").
\item She has never been able to cure her own inability to speak.
\item Include a wise old raven who follows her everywhere.
\item The raven criticizes her recipes.
\item The story must be exactly six sentences long.
\item During the story, reveal that the raven is the villagers' former healer, cursed into that form years ago.
\item End with the witch choosing to break the curse even though it means losing her only companion.
}

\examplecard{Comedy}{The Pop Quiz Panic}{%
Write a comedy story about a kid who realizes mid-quiz that he studied for the wrong chapter, told in first person. He has a habit of tapping his pencil louder when he is panicking. The story must be six sentences long. During the story, reveal that the \lq\lq wrong chapter\rq\rq\ he studied is actually next week's material, so he's somehow more prepared for that quiz than anyone else. Include the exact word \lq\lq doomed\rq\rq\ somewhere in the text. End with the teacher announcing the quiz got postponed to next week anyway.%
}{%
\item Write a comedy story about a kid who realizes mid-quiz that he studied for the wrong chapter.
\item Write the entire story in first person.
\item He has a habit of tapping his pencil louder the more panicked he gets.
\item Include the exact word \lq\lq doomed\rq\rq\ somewhere in the text.
\item The story must be exactly six sentences long.
\item During the story, reveal that the \lq\lq wrong chapter\rq\rq\ he studied is actually next week's material, so he's somehow more prepared for that quiz than anyone else.
\item End with the teacher announcing the quiz got postponed to next week anyway.
}

\end{document}